\documentclass{article}

\usepackage[preprint]{neurips_2024}

\usepackage[utf8]{inputenc} 
\usepackage[T1]{fontenc}    
\usepackage{hyperref}       
\usepackage{url}            
\usepackage{booktabs}       
\usepackage{amsfonts}       
\usepackage{nicefrac}       
\usepackage{microtype}      
\usepackage{xcolor}         

\usepackage{hyperref}
\hypersetup{
 colorlinks=true,
 urlcolor=cyan,
}

\title{Expanding Foundational Language Capabilities in Open-Source LLMs through a Korean Case Study}

\author{
  Moreh\thanks{A detailed contributor list can be found in the appendix of this paper.}
}

\begin{document}

\maketitle

\begin{abstract}
  We introduce Llama-3-Motif, a language model consisting of 102 billion parameters, specifically designed to enhance Korean capabilities while retaining strong performance in English. Developed on the Llama 3 architecture, Llama-3-Motif employs advanced training techniques, including LlamaPro and Masked Structure Growth, to effectively scale the model without altering its core Transformer architecture. Using the MoAI platform for efficient training across hyperscale GPU clusters, we optimized Llama-3-Motif using a carefully curated dataset that maintains a balanced ratio of Korean and English data. Llama-3-Motif shows decent performance on Korean-specific benchmarks, outperforming existing models and achieving results comparable to GPT-4.
  
\end{abstract}

\section{Introduction}

The rapid advancement of large language models (LLMs) has significantly transformed the field of natural language processing. However, there are still notable performance gaps for languages like Korean, primarily due to the scarcity of high-quality datasets and the computational challenges involved in scaling models efficiently. To address these issues, we present \emph{Llama-3-Motif}, a 102 billion-parameter language model designed to enhance the proficiency in Korean while maintaining robust performance in English. Llama-3-Motif builds on the Llama 3~\cite{llama3} pre-trained model, incorporating advanced training methodologies. Specifically, we employ techniques such as LlamaPro~\cite{llamapro} for depth expansion and Masked Structure Growth (MSG)~\cite{msg} for width expansion, allowing for scalable model growth without altering the core Transformer architecture~\cite{attention}. These approaches aim to close the performance gap in Korean language tasks while preserving the overall linguistic capabilities of the model.

A key component of our development process is the use of \emph{MoAI Platform}\footnote{\url{https://moreh.io/product}}, an advanced AI infrastructure designed to streamline the training of large-scale models. This platform manages thousands of GPU clusters, offering features such as automatic parallelization, GPU virtualization, and dynamic GPU allocation. Using these capabilities, the MoAI platform accelerates experiments and optimizations, enabling extensive tasks such as hyperparameter tuning and adjustments for alignment without the burden of manual GPU management. This efficiency allows us to concentrate on refining Llama-3-Motif's architecture and performance.

We curate a comprehensive high-quality dataset of approximately 194 billion tokens, carefully balanced between Korean and English content to improve language proficiency. Through rigorous data collection and filtering processes, we ensure the dataset's relevance and quality, enabling the model to tackle complex, domain-specific tasks in Korean. Following pretraining, we fine-tune and align the model using advanced techniques such as Noisy Embedding Instruction Fine Tuning (NEFTune)~\cite{neftune} and Kahneman-Tversky Optimization (KTO)~\cite{kto}, significantly improving its efficacy and generalization performance.

Evaluations using benchmarks such as the Korean Multitask Language Understanding (KMMLU)~\cite{kmmlu} metric and the KorMedMCQA~\cite{kormedmcqa} data set demonstrate that Llama-3-Motif surpasses existing Korean specialized models by 9- 40\% / 4.9-16.6 points, while also achieving competitive results compared to GPT-4~\cite{gpt4}.

Our contributions include the following.
\begin{itemize}
    \item \textbf{Dataset Curation}: Developing a high-quality, balanced Korean and English dataset to enhance bilingual language proficiency.
    \item \textbf{Model Scaling}: Proposing an efficient methodology to scale a language model from 70 billion to 102 billion parameters through progressive training techniques.
    \item \textbf{Advanced Fine-Tuning Techniques}: Utilizing NEFTune and KTO in post-training to enhance the model's performance.
    \item \textbf{Performance Evaluation}: Demonstrating Llama-3-Motif's decent performance on specialized benchmarks for Korean language tasks.
\end{itemize}

Through this work, we aim to close the performance gap in Korean language models while providing valuable insights into efficient model scaling and the use of advanced AI infrastructure for less-resourced languages.

\section{Model architecture and scaling}

Llama-3-Motif utilizes a standard dense Transformer architecture~\cite{attention}, with the Llama 3 70B~\cite{llama3} pretrained model as its base. A key contributor to our performance improvements is the increased scale of the model.

The expansion of our model was achieved through a progressive training approach. This methodology facilitates the efficient scaling of the existing model by leveraging its current weights while maintaining the original Transformer architecture and enhancing performance. The adoption of this progressive training strategy represents a crucial design decision aimed at improving pre-trained Transformer models. In this context, we explored two primary options: expanding the model's depth and its width.

The "depth expansion" refers to increasing the number of layers, while "width expansion" involves enlarging the hidden size and intermediate size of various components, such as RMS normalization layers, feedforward layers, embedding layers, and attention layers.

To determine the most effective approach for expansion, we conducted four preliminary experiments using Qwen 1.8B~\cite{qwen} as the base model. In these trials, we focused exclusively on increasing the depth by 50\%, without implementing any width expansion.

The experiments employed several methodological approaches, including:
\begin{itemize}
    \item Initialization of a 2.7 billion parameter model, expanded from Qwen 1.8B, based on a normal distribution derived from the base model's parameters.
    \item Staged expansion~\cite{staged}
    \item Implementation of LlamaPro~\cite{llamapro}
    \item Depth-up scaling~\cite{solar}
\end{itemize}

The selection of LlamaPro for depth expansion was based on our preliminary results. For width expansion, we employed MSG. The number of layers was increased by 20\%, while the head dimension was preserved. Adjustments were made to the hidden and intermediate sizes to accommodate these changes. Table~\ref{motif-configuration} presents the architecture of Llama-3-Motif.

\begin{table}
  \caption{Overview of Llama-3-Motif 102B}
  \label{motif-configuration}
  \centering
  \begin{tabular}{ll}
    \toprule
    &  Llama-3-Motif-102B      \\
    \midrule
    Layers & 96                \\
    Model Dimension & 9,216    \\
    FFN Dimension & 30,720     \\
    Attention Heads & 72       \\
    Key/Value Heads & 8        \\
    Vocabulary Size & 128,000  \\
    \bottomrule
  \end{tabular}
\end{table}

Thus, the expanded model was initialized by implementing LlamaPro’s depth expansion on the base model, followed by the width expansion facilitated by MSG which utilizes a masking mechanism to maintain the capabilities of the initial model by neutralizing the effects of the newly introduced neurons. This approach allows for a gradual emphasis on the importance of these neurons throughout the training process. Initially, the newly added parameters are masked, and they are progressively unmasked at each training step for every layer.

\section{Infrastructure}

We employed the MoAI Platform(\url{https://moreh.io/product}) to train Llama-3-Motif. The MoAI Platform is an AI infrastructure specifically designed to facilitate the development of large-scale deep learning models by managing thousands of GPU clusters required for both training and inference. Key features of the platform include automatic parallelization, GPU virtualization, and dynamic GPU allocation, all of which collectively enhance the efficiency of the model development process.

In our work with Llama-3-Motif, we conducted extensive experiments that included hyperparameter tuning, alignment adjustment methods, exploration of scaling techniques, and NEFT embedding~\cite{neftune}, using MoAI Platform with hundreds of AMD MI250 GPUs. These iterative experiments were essential for identifying the optimal scaling and training techniques for the better model. In particular, the platform's automatic parallelization feature proved invaluable, enabling us to efficiently conduct multiple experiments without much considerations on manual configuration of the GPU infrastructure or concerns regarding efficient/effective resource allocation. By treating numerous GPUs as a single virtual device, we were able to concentrate solely on refining the training methods for the Llama-3-Motif model, thereby significantly streamlining the development process and accelerating our progress toward better model performance.

\section{Pre-training}

To tailor the model with expanded parameters for specific capabilities, additional pre-training is necessary. In particular, we identify the target domain data as a critical factor in enhancing performance on downstream tasks.

The continual pre-training dataset utilized for our model consists of an extensive corpus of approximately 194 billion tokens. This dataset was meticulously curated to achieve a balance between Korean and English language proficiency, with a strong emphasis on Korean data to support our primary objective of enhancing the model's proficiency in Korean.

\subsection{Data composition}

The dataset maintained a carefully calibrated ratio of approximately 9:1 between Korean and English content. This distribution was strategically designed to achieve two primary objectives:
\begin{itemize}
    \item To prioritize the enhancement of the model’s Korean language capabilities through comprehensive exposure to a diverse range of Korean texts.
    \item To concurrently preserve the model’s proficiency in English, ensuring its effectiveness in bilingual or English only contexts.
\end{itemize}

\subsection{Data collection}

The majority of the Korean dataset was sourced from web-crawled documents. We aggregated 194 billion tokens of data from a wide range of sources, including news articles, blog posts, and professional documents such as patents, academic papers, and publicly accessible research reports. Raw data was extracted from various formats, including web pages and different document types. To isolate relevant content from these diverse formats, we applied sophisticated text extraction techniques, followed by advanced filtering strategies to refine and cleanse the dataset.

\subsection{Data filtering}

A comprehensive data processing, filtering, and deduplication pipeline was applied to ensure the dataset's quality and relevance. This rigorous procedure led to a considerable reduction in the number of samples, specifically by 83.59\%. However, the overall volume of text decreased by a relatively smaller margin of 40.13\%. This difference suggests that the filtering process was particularly effective in removing shorter and less informative samples, while retaining more substantial and valuable content.

\section{Post-training}

In machine learning, the quality of the datasets plays a critical role in improving the accuracy, efficiency, and generalizability of the models. However, the collection of fine-grained datasets remains a significant challenge, particularly for non-English languages. To overcome this limitation and develop a high-quality, fine-grained Korean-specific dataset, we implemented a comprehensive data cleaning process on open-source instruction datasets. This process used an internal methodology, LLM as a Judge~\cite{judging} and was further supported by extensive evaluations by human validaters.

In the Supervised Fine-Tuning (SFT) process, we used NEFTune~\cite{neftune} to optimize the performance of SFT with noisy embeddings using NEFT-alpha of 8.

Preference optimization (PO) plays a vital role in facilitating controllable enhancements in large-language models. However, current PO methods have limitations that make them inadequate for application to our Korean language model. Although Proximal Policy Optimization (PPO)~\cite{ppo} has shown promising results in various contexts, its considerable memory demands and the additional training costs associated with the policy model make it suboptimal for certain training scenarios. Likewise, Direct Policy Optimization (DPO)~\cite{dpo} not only requires more memory resources than SFT, but also incurs additional costs related to paired data collection. As a result, DPO may be less practical for real-world applications that rely on hard-to-collect language datasets, such as those in Korean.

To reduce the costs involved in collecting fine-grained, language-specific pairwise datasets, we utilized Kahneman-Tversky Optimization (KTO)~\cite{kto} as an alternative to DPO and PPO, which presumably maintains comparable performance. KTO uses unpaired preference data with a binary signal to learn whether an output is desirable or undesirable for an input, unlike DPO that learns them in pairs, which makes data collection more feasible.

To ensure comparable performances between the aforementioned methods, we conducted evaluations using both LLM-as-a-Judge and human assessment. The results of this comprehensive evaluations confirmed the effectiveness of KTO, leading to its adoption as the human alignment approach.

From our experiments, we observed that naïve hyperparameter tuning with KTO resulted in a notable increase in toxicity. To mitigate this issue, we cached the policy model logits across multiple rounds of hyperparameter tuning, a crucial step in maximizing computational efficiency during preference alignment while conducting several iterations. The training parameters for KTO were configured as follows: a batch size of 128, a learning rate of 1e-6, a NEFT alpha of 0, and KTO lambda values set to Desired: 1.375 and Undesired: 1.

\section{Model evaluation}

We use the Korean Multitask Language Understanding (KMMLU)~\cite{kmmlu} and KorMedMCQA~\cite{kormedmcqa} benchmarks to evaluate Llama-3-Motif's proficiency in Korean. These benchmarks are widely recognized for assessing the performance of Korean language models, each offering distinct insights into comprehension across various domains. KMMLU assesses the general knowledge and reasoning skills of the model in a broad spectrum of Korean subjects, including the humanities, social sciences and natural sciences. KorMedMCQA evaluates the model’s understanding of specialized knowledge within the medical domain through multiple-choice questions in Korean.

Using these benchmarks, we can thoroughly assess Llama-3-Motif’s capabilities in both \emph{general language comprehension in Korean } and \emph{domain-specific expertise}, particularly in professional and technical contexts.

Moreover, as we are developing a medical consultation service powered by Llama-3-Motif, the inclusion of KorMedMCQA is especially relevant, aligning with our focus on assessing the model's proficiency in the medical field.

\subsection{KMMLU}

The KMMLU metric is a comprehensive benchmark designed to evaluate the knowledge acquisition capabilities of language models, particularly those trained on Korean data. KMMLU consists of a diverse set of 35,030 questions across 45 distinct subject areas, sourced from various Korean standardized assessments, including the Public Service Aptitude Test (PSAT), professional certification exams, and the College Scholastic Ability Test (CSAT). These questions cover a wide range of difficulty levels, from high school to expert, allowing for a nuanced assessment of the model’s performance.

We conducted our evaluation using the 5-shot approach whose result is summarized in Table~\ref{kmmlu-table}. The strong performance of our model on the KMMLU benchmark can be largely attributed to the composition of our training dataset. In addition to conventional sources such as blog posts and news articles, the dataset included a substantial proportion of specialized documents, such as domestic academic papers, research reports, and patents. The diverse and professionally-oriented nature of this training corpus has significantly improved the model’s ability to exhibit comprehensive knowledge across the diverse domains.

\begin{table}
  \caption{KMMLU evaluation}
  \label{kmmlu-table}
  \centering
  \begin{tabular}{ll}
    \toprule
    Model                                      &  KMMLU-direct score (5-shot)                                     \\
    \midrule
    Exaone-3.0$^+$                              &  44.5~\cite{exaone}                                             \\
    Solar-10.7B                                 &  41.65~\cite{hyperclovax}                                       \\
    Llama-3-70B-Instruct                        &  54.5$^\dag$                                                    \\
    Llama-3.1-70B-Instruct                      &  52.1$^\dag$                                                    \\
    Qwen1.5-72B                                 &  52.6$^\ddag$                                                   \\
    Qwen1.5-110B                                &  57.45$^\ddag$                                                  \\
    Qwen2-72B-Instruct                          &  64.1$^\dag$                                                    \\
    GPT-4-0125-preview                          &  59.95$^\dag$                                                   \\
    GPT-4o-2024-05-13                           &  64.11$^\ddag$                                                  \\
    Gemini Pro                                  &  50.18~\cite{kmmlu}                                             \\
    Hyperclova X-Large$^+$                      &  53.4~\cite{hyperclovax}                                        \\
    \textbf{Llama-3-Motif-102B}$^+$             &  \textbf{64.74}                                                 \\
    \bottomrule
    \multicolumn{2}{l}{\scriptsize $\dag$: Community report~\cite{kmmluCommunity}, $\ddag$: Measured by the authors.}  \\
    \multicolumn{2}{l}{\scriptsize $+$: Specialized in Korean}
  \end{tabular}
\end{table}

\subsection{KorMedMCQA}

KorMedMCQA is a multiple-choice question-answer dataset based on the Korean medical licensing exams, consisting of questions for doctors, nurses, and pharmacists from 2012 to the present. The data was collected by crawling publicly available questions provided by the Korea Health Personnel Licensing Examination Institute. For the Doctor exam, the dataset includes subjects such as Healthcare Laws and Regulations, General Medicine, and Specialized Medicine, covering a wide range of medical topics. This dataset reflects the regulations, standards, and practices of the Korean healthcare system and serves as a valuable benchmark for assessing the understanding and capabilities of medical AI models within the context of Korean healthcare.


We evaluated various models using 5-shot prompting, calculating both subject-specific scores and overall average scores, and compared the results, as presented in Table~\ref{kormedmcqa-table}.

\begin{table}
  \caption{KorMedMCQA evaluation}
  \label{kormedmcqa-table}
  \centering
  \begin{tabular}{lll}
    \toprule
    Model                          & Doctor (5-shot) & Average (5-shot) \\
    \midrule
    Llama-3.1-70B-Instruct$^\ddag$ & 71.58          & 79.06           \\
    Qwen-1.5-110B-Instruct$^\ddag$ & 48.42          & 65.32           \\
    GPT-4-base-0125$^\dag$         & 76.49          & 83.06  \\
    Llama-3-Motif-102B             & \textbf{77.19} & \textbf{83.34}           \\
    \bottomrule
    \multicolumn{3}{l}{\scriptsize $\dag$: Report from~\cite{kormedmcqa}, $\ddag$: Measured by the authors.}
  \end{tabular}
\end{table}

\section{Conclusion}

In this report, we presented Llama-3-Motif, an advanced language model specifically designed to improve Korean language processing. Llama-3-Motif has demonstrated significant improvements in both general linguistic understanding and specialized domain knowledge, outperforming the capabilities of GPT-4’s base model.

This work represents a step toward reducing the performance gap for languages with limited resources. It also provides valuable insights that can be leveraged in the development of similar models for other under-resourced languages.

\medskip

{
\small
\bibliographystyle{plain}
\bibliography{motifRefs}
}


\appendix

\section{Appendix}

\subsection{Contributions}

All authors sorted alphabetically by last name.

\emph{Technical and management leadership}: Gangwon Jo, Sungmin Lee, Junghwan Lim, Jiyoung Park

\emph{Core contributors}: Dongseok Kim, Jihwan Kim, Junhyeok Lee

\emph{Contributors}: Wai Ting Cheung, Dahye Choi, Kibong Choi, Jaeyeon Huh, Beomgyu Kim, Jangwoong Kim, Taehyun Kim, Haesol Lee, Jeesoo Lee, Dongpin Oh, Changseok Song, Daewon Suh



\end{document}